\documentclass[conference]{IEEEtran}
\usepackage{mathtools}
\usepackage[ruled,vlined]{algorithm2e}
\usepackage{enumitem}
\usepackage{booktabs}
\usepackage{multirow}
\usepackage{siunitx}
\usepackage{graphicx}
\usepackage{xcolor}
\usepackage{cite}
\usepackage{amsmath}
\usepackage{float}
\usepackage{url}
\usepackage[caption=false,font=footnotesize]{subfig}
\usepackage{listings}
\usepackage{tikz}
\usepackage[hidelinks]{hyperref}
\IEEEoverridecommandlockouts
\IEEEpubid{\makebox[\columnwidth]{979-8-3315-5021-9/26/\$31.00~\copyright2026~IEEE \hfill}%
\hspace{\columnsep}\makebox[\columnwidth]{ }}
\usetikzlibrary{shapes.geometric, arrows.meta, positioning}
\begin{document}

\title{TAMS: Task-Aware Multi-View Adaptive Streaming for Wireless Telerobotic Manipulation}

\author{%
\begin{minipage}{\textwidth}

\begin{minipage}[t]{0.32\textwidth}
    \centering
    \normalsize Zexin Deng\\[3pt]
    \small \textit{School of Engineering}\\
    \textit{University of Warwick}\\
    Coventry, United Kingdom\\
    zexin.deng@warwick.ac.uk
\end{minipage}%
\hfill
\begin{minipage}[t]{0.32\textwidth}
    \centering
    \normalsize Zhenhui Yuan\\[3pt]
    \small \textit{School of Engineering}\\
    \textit{University of Warwick}\\
    Coventry, United Kingdom\\
    zhenhui.yuan@warwick.ac.uk
\end{minipage}%
\hfill
\begin{minipage}[t]{0.32\textwidth}
    \centering
    \normalsize Lu Tian\\[3pt]
    \small \textit{School of Engineering}\\
    \textit{University of Warwick}\\
    Coventry, United Kingdom\\
    lu.tian.2@warwick.ac.uk
\end{minipage}

\vspace{1.5em} 

\centerline{%
    \begin{minipage}[t]{0.45\textwidth}
        \centering
        \normalsize Subhash Lakshminarayana\\[3pt]
        \small \textit{School of Engineering}\\
        \textit{University of Warwick}\\
        Coventry, United Kingdom\\
        Subhash.Lakshminarayana@warwick.ac.uk
    \end{minipage}%
    \hspace{1.5em}
\begin{minipage}[t]{0.45\textwidth}
    \centering
    \normalsize Longhao Zou\\[3pt]
    \small \textit{Pengcheng Laboratory \&}\\
    \textit{Southern Univ. of Sci. \& Tech.}\\
    Shenzhen, China\\
    zoulh@pcl.ac.cn
\end{minipage}%
}

\end{minipage}
}

\maketitle
\begin{abstract}
Wireless telerobotic manipulation relies on timely multi-view video feedback, but
the available uplink bandwidth is often limited and dynamic. This paper presents
Task-Aware Multi-View Adaptive Streaming (TAMS), a system that allocates video
bitrate according to the current manipulation phase. TAMS infers task phase from
lightweight robot-side signals and prioritizes the camera view most relevant to
the operator while preserving baseline visibility for secondary views.
Experiments on a six-degree-of-freedom (6-DoF) teleoperation testbed under three constrained network conditions show that TAMS improves primary view Structural
Similarity Index (SSIM), reduces task completion time, and increases trial success
rate compared with equal and static allocation baselines. Under the most constrained bandwidth condition, TAMS reduces mean completion time from $68.9$~s to
$43.9$~s relative to equal allocation and increases trial success rate from
$48$\% to $71$\%. Code is available at: \url{https://github.com/Dzxx623/TAMS}.
\end{abstract}


\section{Introduction}

Telerobotics enables a human operator to control a robotic system across physical
distance, extending human perception, judgment, and manipulation capability
to environments that are difficult, dangerous, or impractical for direct human
access~\cite{niemeyer2008telerobotics}. Representative applications include
disaster response, industrial inspection, nuclear maintenance, space operations,
and other remote intervention scenarios~\cite{hong2018rescuerobot}. In
telerobotic manipulation, the operator must close the loop between remote visual
feedback and local control commands while estimating object pose, gripper
alignment, contact events, and task progress from transmitted video. The quality
and timeliness of visual feedback therefore have a direct influence on
manipulation accuracy, task efficiency, and operator workload.

Practical wireless telerobotic manipulation is constrained by limited and dynamic
uplink resources. In the mobile robot use cases specified by 3GPP TS~22.104,
remote control with video feedback and real-time video streaming are treated as service
requirements for cyber-physical control applications, with a communication
service availability target above 99.9999\% and an end-to-end latency range of
40--500~ms for mobile robot remote control with video feedback~\cite{3gpp22104}. The challenge is amplified when multiple task-relevant
cameras must share the same wireless uplink under bandwidth fluctuation, latency
variation, and packet impairments~\cite{black2024latency}. Recent work has explored haptic-enhanced teleoperation~\cite{huang2025multisensory}, while TeleSim and VISTA provide network-aware benchmarks for telerobotic systems and real-time teleoperation video streaming~\cite{deng2025telesim,deng2026vista}. These efforts motivate adaptive streaming policies that account for both network conditions and task-level teleoperation performance.

Although adaptive bitrate streaming can react to network variation, conventional
policies are usually driven by network performance measurements such as
throughput, delay, and packet loss. They do not explicitly account for what the
operator is trying to accomplish at a given manipulation phase. This is limiting
in multi-view teleoperation, where the utility of each camera view is task-dependent. Global views are more useful for target localization, workspace
awareness, coarse motion guidance, and collision avoidance, whereas a close-up
end-effector view is more useful for gripper alignment, grasping, and release.
Consequently, existing streaming policies leave two issues unresolved: (1)
bitrate adaptation is rarely guided by the operator's current manipulation
objective; and (2) uplink resources are rarely distributed according to the task-dependent utility of each camera view.

To address this gap, we present \textit{TAMS} (\textbf{T}ask-\textbf{A}ware
\textbf{M}ulti-view Adaptive \textbf{S}treaming), a real-time multi-view
streaming system for wireless telerobotic manipulation under dynamic uplink
constraints. \textit{TAMS} infers the current manipulation phase from lightweight
robot-side signals, including end-effector kinematics and gripper events, and
allocates bitrate according to the view utility of that phase. The policy
increases the bitrate share of the view most relevant to the current phase while
maintaining a minimum bitrate for secondary views, improving fine visual detail
without eliminating global situational awareness. Unlike methods that rely on
semantic analysis of video frames, \textit{TAMS} uses compact robot-state signals
and a fixed mapping between task phases and camera views calibrated offline. This
makes the runtime policy lightweight, interpretable, and suitable for deployment
on resource-constrained robotic platforms.

The main contributions of this paper are as follows:
\begin{enumerate}
  \item \textbf{A task-aware formulation of multi-view bitrate allocation:}
  We formulate wireless telerobotic video streaming as a task-dependent allocation
  problem in which camera view importance changes across manipulation phases. The
  formulation captures the need to improve the most useful view under dynamic
  uplink constraints while preserving minimum visibility for the remaining views.

  \item \textbf{A lightweight real-time streaming system for telerobotic manipulation:}
  We design and implement \textit{TAMS}, which combines online phase inference
  from end-effector kinematics and gripper events with bitrate allocation over
  three camera streams. The system avoids semantic analysis on video frames and
  maps directly to encoder bitrate control.

  \item \textbf{A human-in-the-loop evaluation under constrained wireless conditions:}
  We evaluate \textit{TAMS} on a 6-DoF telerobotic manipulation testbed under
  three representative constrained uplink conditions, demonstrating improvements
  in the SSIM of the primary view, task completion time, and trial success rate.
\end{enumerate}

\section{System Model and Problem Formulation}
\label{sec:problem}

We consider a wireless telerobotic manipulation system in which a robot-side
workstation transmits multiple camera streams to a remote operator over a shared
uplink. The operator performs a pick-and-place task consisting of six phases:
\emph{Reach}, \emph{Align}, \emph{Grasp}, \emph{Transport}, \emph{Pre-release},
and \emph{Release}. The camera views provide complementary information: global
views support workspace awareness and coarse motion guidance, while close-range
views support gripper alignment, contact observation, and placement confirmation.
Thus, the value of each view changes with the manipulation phase, and a fixed or
equal-priority allocation may be inefficient under limited bandwidth.

Let $\mathcal{V}=\{1,\dots,V\}$ denote the set of camera streams and
$\mathcal{P}=\{1,\dots,P\}$ the set of manipulation phases. At time $t$, the
system is in phase $s_t \in \mathcal{P}$, and the available uplink budget is
$B_t$. To encode phase-specific view importance, we define a normalized view
weight matrix $W_{p,v}$, where $W_{p,v}\in[0,1]$ and
$\sum_{v\in\mathcal{V}} W_{p,v}=1$ for each phase $p$. The quantity $W_{p,v}$
specifies the desired bitrate share of view $v$ in phase $p$.

The objective is to compute target stream bitrates
$\{b_{v,t}\}_{v\in\mathcal{V}}$ such that
\begin{equation}
\sum_{v\in\mathcal{V}} b_{v,t}\leq B_t,
\end{equation}
while preserving the visual quality of the most important view in the current
phase. A uniform allocation may waste bitrate on less informative streams during
fine manipulation, whereas an overly concentrated allocation may degrade scene
awareness. We therefore seek a real-time allocation policy that adapts to the
current phase, prioritizes the most important stream, and maintains sufficient
visual context from the remaining streams.

\section{TAMS Design}
\label{sec:tams}

\begin{figure}[t]
  \vspace{-4mm}
  \centering
  \includegraphics[
    width=\columnwidth,
    keepaspectratio,
    trim=0pt 10pt 0pt 5pt,
    clip
  ]{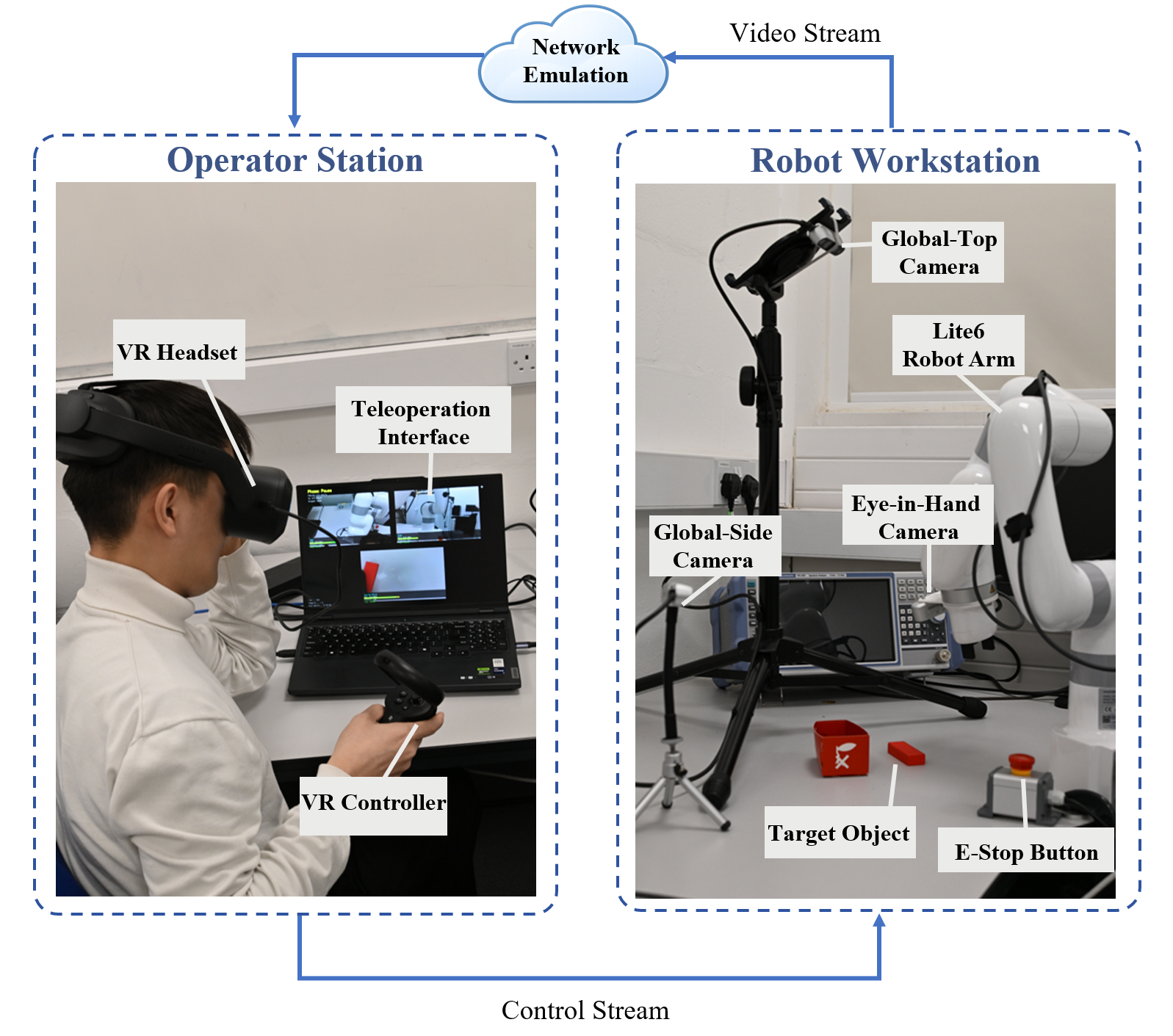}
  \vspace{-6mm}
  \caption{TAMS platform: operator station and VR controller (left), Lite~6 robot
and three-camera setup (right), and emulated uplink (top).}
  \label{fig:system_arch}
  \vspace{-2mm}
\end{figure}

\subsection{Overview of TAMS}

TAMS is a real-time task-aware multi-view adaptive streaming system for wireless
telerobotic manipulation. It allocates bitrate according to the available uplink
budget and the current manipulation phase, prioritizing the most useful view while
retaining baseline visibility for secondary views.

The human-in-the-loop teleoperation platform used to implement and evaluate
TAMS is shown in Fig.~\ref{fig:system_arch}. The robot side consists of a UFACTORY Lite~6
manipulator equipped with a two-finger electric gripper and three camera views:
an Eye-in-Hand (EIH) camera mounted near the end-effector, a Global-Top (GT)
camera, and a Global-Side (GS) camera. 

The operator side uses an HTC VIVE Focus~3 headset and a tracked
six-degree-of-freedom (6-DoF) controller. The headset provides immersive visual
feedback, while the controller captures operator motion and button commands. The
controller pose and button states are sampled at 120~Hz through the OpenVR API.
The host workstation runs Robot Operating System~2 (ROS~2)~\cite{macenski2022ros2}, the robot control
interface, the phase inference module, the video streaming modules, and the VR
interface. Video decoding is accelerated by an NVIDIA RTX~4060 GPU. The control
channel remains separate from the video uplink so that the streaming policy
affects only visual feedback.

The tracked controller is used as a motion input device for Cartesian
teleoperation. At each control step, TAMS estimates translational and rotational
controller velocities from successive tracked poses, smooths them using exponential
moving averages, transforms them into the robot control frame using a calibrated
mapping matrix, and applies velocity limits and workspace bounds before sending
Cartesian servo commands to the Lite~6 manipulator. This implementation follows
the general design motivation of prior work showing the suitability of
motion-tracked 6-DoF input for robot
teleoperation~\cite{rakita2017motion,whitney2019comparing}, while using direct
Cartesian servo control rather than a RelaxedIK-based motion-synthesis backend.

The controller trigger acts as a deadman switch and clutching mechanism for
motion control. The grip button switches between translation and orientation
control, while the A and B buttons command gripper closing and opening.

At runtime, TAMS operates through three coupled stages. First, it infers the
current manipulation phase from robot-side kinematics and gripper events. Second,
it obtains the phase-specific view weights from an offline-calibrated view weight
matrix indexed by phase and view. Third, it allocates the available uplink budget
across the EIH, GT, and GS streams according to the inferred phase and the
corresponding view weights.

Each camera view is implemented as an independent GStreamer
pipeline.\footnote{GStreamer: \url{https://gstreamer.freedesktop.org/}}
In each sender pipeline, frames are injected through \texttt{appsrc}, converted,
encoded by \texttt{x264enc}~\cite{x264}, packetized into Real-time Transport Protocol (RTP)
payloads, and sent over User Datagram Protocol (UDP):
\begin{center}
\small\ttfamily
appsrc $\rightarrow$ videoconvert $\rightarrow$ x264enc\\
$\rightarrow$ h264parse $\rightarrow$ rtph264pay $\rightarrow$ udpsink
\end{center}
The receiver performs the reverse operations:
\begin{center}
\small\ttfamily
udpsrc $\rightarrow$ rtpjitterbuffer $\rightarrow$ rtph264depay\\
$\rightarrow$ avdec\_h264 $\rightarrow$ videoconvert $\rightarrow$ autovideosink
\end{center}
This structure is instantiated separately for the EIH, GT, and GS views. All
streams use the same frame rate, resolution, transport protocol, and decoder
configuration, while TAMS updates the H.264 target bitrate of each stream
independently. Thus, bitrate allocation maps directly to encoder control rather
than to a jointly encoded composite video.

\subsection{Task Phase Inference}
\label{sec:phase_inference}

The first runtime component of TAMS is the online task phase inference module.
The module estimates the current manipulation phase using lightweight robot-side
signals that are already available from the robot control loop. Specifically, it
uses the smoothed end-effector speed $\bar{v}_t$, the smoothed vertical speed
$\bar{v}_{z,t}$, and the binary gripper state $g_t\in\{0,1\}$, where 0 denotes an
open gripper and 1 denotes a closed gripper.

The pick-and-place task is divided into six manipulation phases:
\emph{Reach}, \emph{Align}, \emph{Grasp}, \emph{Transport},
\emph{Pre-release}, and \emph{Release}. Gripper-state transitions provide
high-confidence event cues for contact-related phases. An open-to-closed
transition indicates \textsc{Grasp}, while a closed-to-open transition indicates
\textsc{Release}. When no gripper transition occurs, velocity-based conditions
distinguish coarse motion phases from fine manipulation phases. Sustained
downward motion with an open gripper indicates \textsc{Align}, whereas sustained
downward motion with a closed gripper indicates \textsc{Pre-release}. Otherwise,
the system assigns \textsc{Reach} when the gripper is open and \textsc{Transport}
when the gripper is closed.

Short operator pauses may occur within or between manipulation phases. TAMS does
not treat these pauses as a separate phase. Instead, when the smoothed
end-effector speed falls below a small threshold, the controller holds the most
recently inferred phase. This prevents oscillatory bitrate reallocation during
temporary stillness. The velocity thresholds and dwell times depend on the platform and were
calibrated on the Lite~6 testbed and kept fixed across all experimental
conditions. Algorithm~\ref{alg:state_transition} summarizes the resulting online
phase inference procedure.

\begin{algorithm}[t]
\small
\SetAlgoLined
\DontPrintSemicolon
\SetKwInOut{Input}{\textbf{Input}}
\SetKwInOut{Output}{\textbf{Output}}
\providecommand{\Phase}[1]{\textsc{#1}}

\caption{Online Task Phase Inference}
\label{alg:state_transition}

\Input{Smoothed speed $\bar{v}_t$, smoothed vertical speed $\bar{v}_{z,t}$,
gripper state $g_t$, previous gripper state $g_{t-1}$, previous phase $s_{t-1}$,
descending-motion timer $\tau_{\mathrm{desc}}$, control period $\Delta t$}
\Output{Updated phase $s_t$}

\BlankLine
\textit{Parameters:} velocity thresholds $v_\epsilon$, $v_z^{\min}$ and dwell
times $T_{\mathrm{align}}$, $T_{\mathrm{pre}}$\;

$\Delta g \leftarrow g_t - g_{t-1}$\;

\If{$\Delta g > 0$}{
    \Return{$\Phase{Grasp}$}\;
}

\If{$\Delta g < 0$}{
    \Return{$\Phase{Release}$}\;
}

\If{$\bar{v}_t \leq v_\epsilon$}{
    \Return{$s_{t-1}$}\;
}

\eIf{$\bar{v}_{z,t} < -v_z^{\min}$}{
    $\tau_{\mathrm{desc}} \leftarrow \tau_{\mathrm{desc}} + \Delta t$\;
}{
    $\tau_{\mathrm{desc}} \leftarrow 0$\;
}

\eIf{$g_t = 0$}{
    \eIf{$\tau_{\mathrm{desc}} \geq T_{\mathrm{align}}$}{
        \Return{$\Phase{Align}$}\;
    }{
        \Return{$\Phase{Reach}$}\;
    }
}{
    \eIf{$\tau_{\mathrm{desc}} \geq T_{\mathrm{pre}}$}{
        \Return{$\Phase{Pre-release}$}\;
    }{
        \Return{$\Phase{Transport}$}\;
    }
}
\end{algorithm}

\subsection{View Weight Determination}
\label{sec:weight_config}

The second component of TAMS is the determination of phase-specific view weights. The online allocator relies on a fixed view weight matrix indexed by
phase and view, $W_{p,v}$, where $p$ denotes the manipulation phase and $v$
denotes the camera view. Each row of the matrix is normalized so that
\begin{equation}
\sum_{v\in\mathcal{V}} W_{p,v}=1.
\end{equation}
The value $W_{p,v}$ represents the desired relative importance of view $v$ during
phase $p$.

This design is motivated by the fact that the utility of each camera view changes
with the manipulation phase. During coarse motion phases, such as \emph{Reach}
and \emph{Transport}, global views are more useful for target localization,
workspace awareness, and collision avoidance. During fine manipulation phases,
such as \emph{Align}, \emph{Grasp}, \emph{Pre-release}, and \emph{Release}, the
EIH stream becomes more important because it provides local visual detail for
gripper alignment, object contact, and placement confirmation. This follows the
broader observation from task-oriented communication that the usefulness of
transmitted information depends on the robot's current task phase~\cite{chen2025goal},
and from multi-camera teleoperation studies that different viewpoints provide
complementary visual cues for remote manipulation~\cite{praveena2022control}.

TAMS obtains $W_{p,v}$ offline through a compact calibration procedure under
unconstrained visual quality. A small calibration sample ($N=8$) performed the
same pick-and-place task with all three views streamed at full quality. Afterward,
participants reviewed clips segmented by task phase, rated the necessity of each
camera view for each task phase, and identified the single most critical view per
phase under an imagined severe bandwidth constraint. These ratings were used only
to calibrate the streaming policy and were not treated as a separate user study
contribution.

The final matrix was constructed using a rule that assigns the largest share to a single dominant view per phase. For each
phase, the view identified as most critical receives the largest bitrate share,
while the remaining views retain nonzero shares to preserve situational awareness.
The resulting phase-specific allocation weights are shown in
Table~\ref{tab:weight-matrix}.

\begin{table}[!t]
  \centering
  \caption{Phase-specific view weights used by TAMS.}
  \label{tab:weight-matrix}
  \small 
  \renewcommand{\arraystretch}{1.1}
  \setlength{\tabcolsep}{8pt} 
  \begin{tabular}{|l|c|c|c|}
    \hline
    \textbf{Phase} & \textbf{EIH} & \textbf{GT} & \textbf{GS} \\
    \hline
    Reach       & 0.23          & \textbf{0.50} & 0.27 \\
    \hline
    Align       & \textbf{0.60} & 0.20          & 0.20 \\
    \hline
    Grasp       & \textbf{0.60} & 0.20          & 0.20 \\
    \hline
    Transport   & 0.25          & 0.30          & \textbf{0.45} \\
    \hline
    Pre-release & \textbf{0.55} & 0.25          & 0.20 \\
    \hline
    Release     & \textbf{0.55} & 0.25          & 0.20 \\
    \hline
  \end{tabular}

\vspace{2mm}
\parbox{0.9\columnwidth}{%
\centering
\scriptsize
\textit{Note:} Rows sum to 1.0; bold values indicate the primary view.
}
\vspace{-2mm}
\end{table}

The matrix is intentionally fixed during runtime. This avoids online user
modelling, reduces computational overhead, and keeps the allocation policy
interpretable. It also separates the estimation of view utility from the
real-time communication controller, which is important for deployment on
resource-constrained robotic platforms.

\subsection{Bitrate Allocation}
\label{sec:bitrate_allocation}

The third component of TAMS is the bitrate allocation module. Once the current
phase $s_t$ has been inferred, the allocator retrieves the available uplink budget
$B_t$, estimated from outgoing RTP sender statistics over a sliding window. It then
computes target bitrates for the three video streams using the row of
$W_{p,v}$ corresponding to the current phase.

Without a minimum floor, the nominal allocation rule is
\begingroup
\setlength{\abovedisplayskip}{4pt}
\setlength{\belowdisplayskip}{4pt}
\begin{equation}
b_{v,t}^{\mathrm{nom}} = W_{s_t,v} B_t,
\label{eq:bandwidth_allocation_nominal}
\end{equation}
\endgroup
where $b_{v,t}^{\mathrm{nom}}$ is the nominal target bitrate of view $v$ at time
$t$. This rule directly couples communication control to manipulation context.
During fine manipulation phases, the EIH stream receives a larger share of the
available budget. During coarse motion phases, bitrate is shifted toward the
global view that provides the most useful spatial context.

To prevent complete loss of secondary views under severe bandwidth pressure, TAMS
enforces a minimum bitrate floor for each stream, denoted by $b_{\min}$. When the
current uplink budget is sufficient to satisfy the floor for all streams, the
allocator first reserves the floor and then distributes the remaining budget
according to the phase-specific weights:
\begingroup
\setlength{\abovedisplayskip}{4pt}
\setlength{\belowdisplayskip}{4pt}
\begin{equation}
b_{v,t}^{\mathrm{target}}
=
b_{\min}
+
W_{s_t,v}\left(B_t - V b_{\min}\right),
\quad
B_t \geq V b_{\min}.
\label{eq:bandwidth_allocation_floor}
\end{equation}
\endgroup
Here, $V$ is the number of video streams. In our implementation, $V=3$ for the
EIH, GT, and GS views. If the available budget falls below $Vb_{\min}$, the floor
cannot be satisfied for all views. In that case, all streams are degraded
proportionally:
\begingroup
\setlength{\abovedisplayskip}{4pt}
\setlength{\belowdisplayskip}{4pt}
\begin{equation}
b_{v,t}^{\mathrm{target}} = \frac{B_t}{V},
\quad
B_t < V b_{\min}.
\label{eq:bandwidth_allocation_collapse}
\end{equation}
\endgroup

The resulting allocation policy has three practical properties. First, it protects
the most important view in the current manipulation phase. Second, it preserves
minimum visibility for secondary views so that the operator does not lose global
situational awareness. Third, it maps directly to independent encoder controls for
the three RTP video streams, allowing TAMS to operate in real time without
requiring semantic analysis of video frames or jointly encoded multi-view video.

\section{Experimental Methodology}
\label{sec:method}

\subsection{Experimental Use Case and Task}

We evaluate TAMS in a wireless telerobotic pick-and-place use case. In each trial,
the operator guided the robot end-effector to a target block, aligned the gripper,
closed the gripper to grasp the block, transported the block to a container, and
released it. This task was selected because it contains both coarse motion phases,
where global spatial context is important, and fine manipulation phases, where
EIH visual detail is critical. Trials were limited to 120~s; success required correct placement within 90~s
and no more than two drops.
The experiment used a $3 \times 3$ within-subjects design with two factors:
\textit{Streaming Strategy} and \textit{Network Condition}. The three streaming
strategies were equal allocation, static allocation, and TAMS. Equal allocation
assigns the same bitrate share to all three views; static allocation assigns fixed weights of $(0.50,0.25,0.25)$ to EIH, GT, and
GS across the whole task; TAMS uses phase-specific view weights. The three network
conditions were S1, S2, and S3, corresponding to progressively more constrained
wireless uplink conditions. Each participant completed all nine conditions with
three repetitions per condition, yielding 432 trials in total. Condition order was
counterbalanced using a Balanced Latin Square.

\subsection{Participants}

Sixteen participants ($N=16$, aged 21--30, $M=24.5$, $SD=3.2$; 10 male, 6 female)
were recruited from a local university community. Eligibility required normal or
corrected-to-normal vision. Most participants were novice teleoperators. None of
the participants took part in the offline weight calibration described in
Section~\ref{sec:weight_config}.

\subsection{Streaming Conditions, Measures, and Analysis}

All strategies used the same capture resolution, frame rate, transport protocol,
and encoder stack, differing only in how the available video bandwidth was
allocated across the three streams. Each camera stream was captured at
640$\times$480 and 30~fps, encoded using H.264, and transmitted over RTP/UDP. The
video bandwidth in Table~\ref{tab:network_params} denotes the compressed H.264
bandwidth shared by the three streams, rather than the uncompressed raw camera data
rate. A safety factor of 0.92 was applied before computing encoder target bitrates.

Network emulation was applied using Linux \texttt{tc} on the video uplink path
only, while the control channel remained unconstrained. Table~\ref{tab:network_params}
summarizes the three network conditions.

\begin{table}[!t]
  \centering
  \caption{Network emulation parameters.}
  \label{tab:network_params}
  \scriptsize
  \renewcommand{\arraystretch}{1.12}
  \setlength{\tabcolsep}{4pt}
  \resizebox{0.98\columnwidth}{!}{%
  \begin{tabular}{|l|c|c|c|}
    \hline
    \textbf{Parameter} & \textbf{S1: High} & \textbf{S2: Medium} & \textbf{S3: Low} \\
    \hline
    One-way latency & 15 ms & 40 ms & 80 ms \\
    \hline
    Jitter & 2 ms & 8 ms & 15 ms \\
    \hline
    Packet loss & 0.01\% & 0.3\% & 1.0\% \\
    \hline
    Video bandwidth & 12 Mbps & 6 Mbps & 3 Mbps \\
    \hline
  \end{tabular}%
  }
  \vspace{-2mm}
\end{table}

\section{Results}
\label{sec:results}

\subsection{Bitrate Tracking}
\label{sec:bt}

\begin{figure}[!t]
\centering
\includegraphics[width=0.90\linewidth]{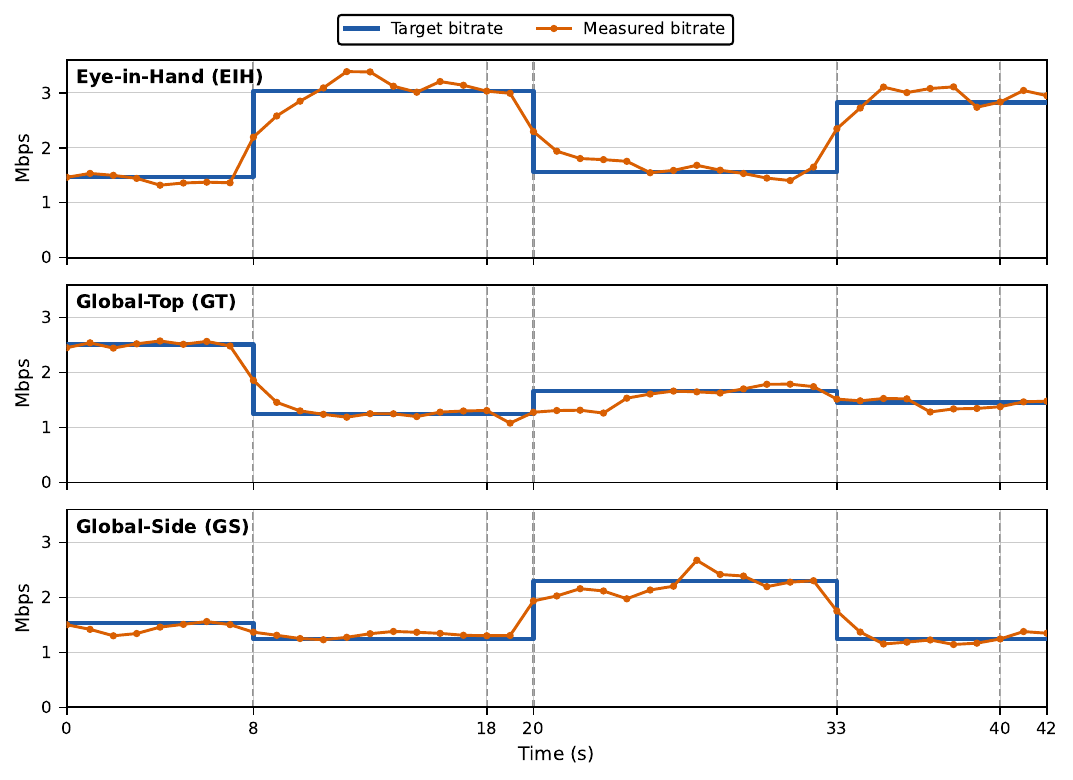}
\vspace{-2mm}
\caption{Target and measured bitrate traces for the three RTP/H.264 streams
during a representative six-phase trial under the 6~Mbps network condition.
Vertical dashed lines indicate task phase transitions.}
\label{fig:bitrate_tracking}
\vspace{-2mm}
\end{figure}

As shown in Fig.~\ref{fig:bitrate_tracking}, the encoder bitrate follows
the targets conditioned on the current phase. Under the 6~Mbps condition, TAMS shifts bitrate to
the EIH stream during \emph{Align}, \emph{Grasp}, \emph{Pre-release}, and
\emph{Release}, and to the global streams during \emph{Reach} and
\emph{Transport}. The measured RTP/UDP traces follow the same pattern, with only
short transients after phase changes. This confirms that the allocation rule is
both computed by the controller and realized by the video pipeline.

\subsection{Visual Fidelity}
\label{sec:vf}

For each received frame, we compute Structural Similarity Index (SSIM) against
the reference frame at the sender and report the SSIM of the \emph{primary view},
defined as the view with the largest phase-specific weight in
Table~\ref{tab:weight-matrix}. SSIM is used because it better reflects
perceptual quality than PSNR~\cite{wang2004ssim}.

\begin{figure*}[t]
  \centering

  \subfloat[S1: 12~Mbps / 15~ms]{%
    \includegraphics[width=0.31\linewidth]{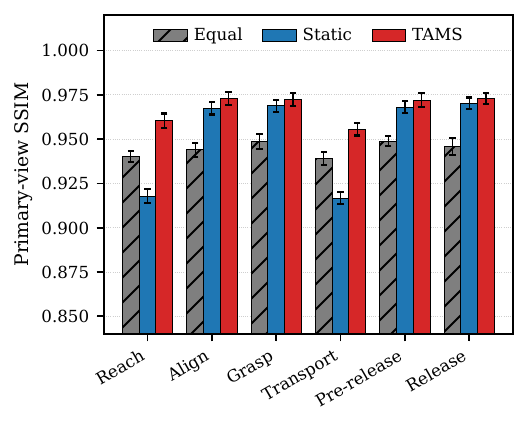}%
    \label{fig:vf_s1}}\hfill
  \subfloat[S2: 6~Mbps / 40~ms]{%
    \includegraphics[width=0.31\linewidth]{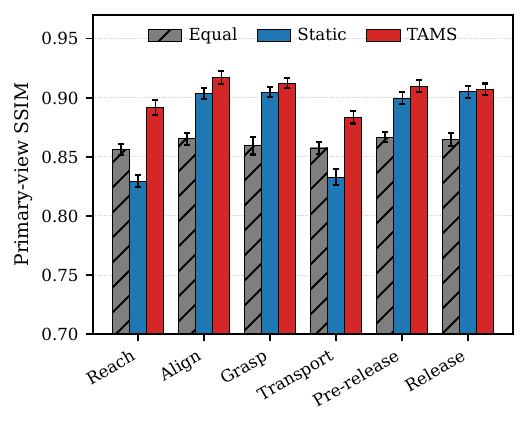}%
    \label{fig:vf_s2}}\hfill
  \subfloat[S3: 3~Mbps / 80~ms]{%
    \includegraphics[width=0.31\linewidth]{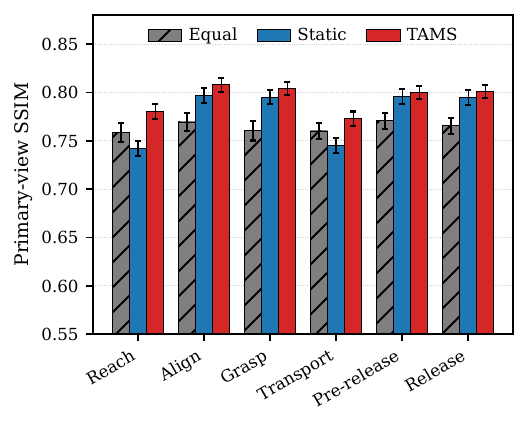}%
    \label{fig:vf_s3}}

  \caption{Primary view SSIM across the six manipulation phases under the three
  network conditions. Error bars show 95\% confidence intervals.}
  \label{fig:fidelity}
\end{figure*}

As shown in Fig.~\ref{fig:fidelity}(a)--(c), TAMS achieves the highest
primary view SSIM in nearly all phase and network combinations. Averaged over the
six phases, TAMS reaches $0.968$, $0.904$, and $0.794$ under S1, S2, and S3,
respectively, compared with $0.951$, $0.879$, $0.778$ for Static and $0.944$,
$0.862$, $0.764$ for Equal.

The advantage becomes more pronounced under S2 and S3, where bandwidth is limited.
Static allocation can degrade performance when the primary view is not the
EIH stream. For example, under S1 during \emph{Reach}, primary view SSIM
is $0.940$ for Equal, $0.918$ for Static, and $0.960$ for TAMS. This shows that
fixed prioritization improves views for fine manipulation at the expense of global context.
TAMS avoids this tradeoff by adapting the primary stream to the manipulation phase.

\subsection{Task Performance}
\label{sec:tp}

We next evaluate whether improvements in visual fidelity translate into better
task outcomes. Completion time is measured from the first deadman switch
activation to the gripper opening event. Trial success is defined using the
criterion in Section~\ref{sec:method}.

\begin{figure}[t]
  \centering
  \includegraphics[width=0.62\linewidth]{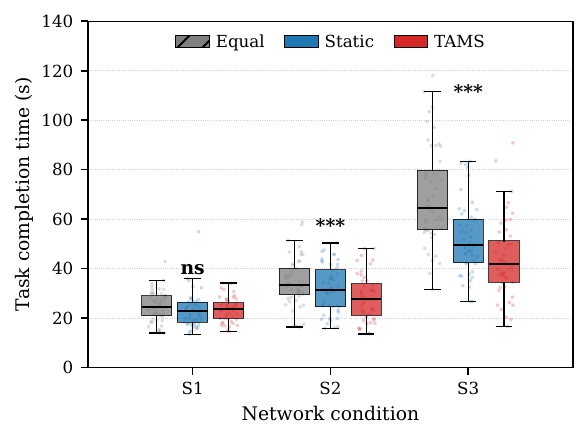}
    \vspace{-4mm}
  \caption{Task completion time across the three streaming strategies and three
network conditions. Box edges denote interquartile range, central line denotes
median, whiskers extend to the 5th--95th percentile, and individual trials are
overlaid as jittered dots. Significance markers indicate the results of Welch's
$t$-test between TAMS and Equal: \texttt{ns} and \texttt{***}~$p<.001$.}
  \label{fig:tp_completion}
\end{figure}

As shown in Fig.~\ref{fig:tp_completion}, completion time increases as the
network degrades, but less so under TAMS. Mean completion times for TAMS,
Static, and Equal are $23.5$, $23.4$, and $24.7$~s under S1; $28.1$, $31.7$, and
$34.8$~s under S2; and $43.9$, $52.1$, and $68.9$~s under S3. A linear
mixed-effects model reveals a significant TAMS$\times$S3 interaction
($\hat{\beta}=-23.47$~s, $\mathrm{SE}=3.22$, $p<.001$). Effect sizes are large
under constrained conditions ($d=-1.38$ for S3, $d=-0.72$ for S2), while S1 shows
no significant difference.

\begin{figure}[t]
  \centering
  \includegraphics[
    width=0.62\linewidth,
    trim=0 5pt 0 0,
    clip
  ]{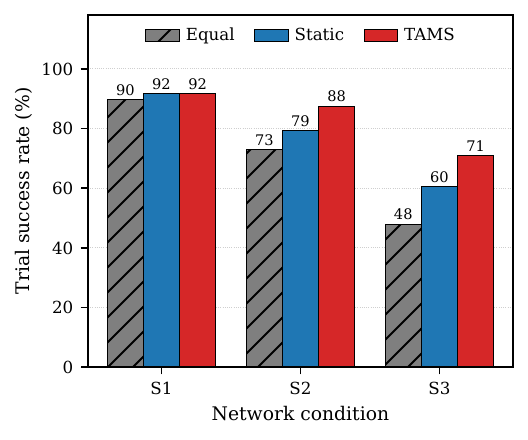}
      \vspace{-4mm}
  \caption{Trial success rate across the three streaming strategies and three
  network conditions. Numeric labels above each bar denote the corresponding
  percentage.}
  \label{fig:tp_success}
\end{figure}

A consistent pattern for success rate is shown in Fig.~\ref{fig:tp_success}. Under
S1, all strategies are near ceiling ($90$--$92$\%). Under S2, success is $73$\%
for Equal, $79$\% for Static, and $88$\% for TAMS. Under S3, the gap widens to
$48$\%, $60$\%, and $71$\%, respectively. TAMS thus improves success by $23$
percentage points over Equal and $11$ over Static under the most constrained
condition.

\section{Conclusion}
\label{sec:conclusion}

This paper presented TAMS, a task-aware multi-view adaptive streaming system for
wireless telerobotic manipulation. By inferring the current manipulation phase
from lightweight robot-side signals, TAMS allocates more bitrate to the camera
view most relevant to the operator while preserving baseline visibility for
secondary views. Experiments on a 6-DoF teleoperation testbed show that TAMS
realizes the intended phase-specific allocation at the streaming layer, improves
primary view SSIM, reduces completion time, and increases trial success rate,
especially under constrained uplink conditions. These results show that the utility of teleoperation video is task dependent. Future work will extend TAMS to online adaptation of view weights, richer task inference, real mobile network traces, and larger user
studies.

\bibliographystyle{IEEEtran}
\bibliography{sample-base}

\end{document}